\documentclass[10pt,twocolumn,letterpaper]{article} 
\usepackage{avss}
\usepackage{epsfig}
\usepackage{times}
\usepackage{amsmath,amsfonts}
\usepackage{algorithm}
\usepackage{array}
\usepackage{subfigure}
\usepackage{textcomp}
\usepackage{stfloats}
\usepackage{url}
\usepackage{verbatim}
\usepackage{graphicx}
\usepackage{svg}
\usepackage{bm} 
\usepackage{times}
\usepackage{epsfig}
\usepackage{cite}
\usepackage[numbers]{natbib}
\usepackage{stfloats}
\usepackage{physics}
\usepackage{amssymb}
\usepackage{gensymb}
\usepackage{algpseudocode}
\usepackage{subcaption}
\usepackage{textcomp}
\usepackage{xcolor}
\usepackage{adjustbox}
\usepackage{multirow}
\usepackage{tabularx}
\usepackage{booktabs}
\usepackage[breaklinks=true,bookmarks=false]{hyperref}
\usepackage{cleveref}
\usepackage{balance}

\avssfinalcopy 

\def\avssPaperID{48} 

\ifavssfinal\fi
\begin{document}
\title{MirrorCalib: Utilizing Human Pose Information for Mirror-based Virtual Camera Calibration}


\author{Longyun Liao\\
McMaster University\\
{\tt\small liaol13@mcmaster.ca}
\and
Rong Zheng\\
McMaster University\\
{\tt\small rzheng@mcmaster.ca}
\and
Andrew Mitchell\\
McMaster University\\
{\tt\small mitchaj2@mcmaster.ca}
}
\maketitle

\begin{abstract}
In this paper, we present the novel task of estimating the extrinsic parameters of a virtual camera relative to a real camera in exercise videos with a mirror. This task poses a significant challenge in scenarios where the views from the real and mirrored cameras have no overlap or share salient features. To address this issue, prior knowledge of a human body and 2D joint locations are utilized to estimate the camera extrinsic parameters when a person is in front of a mirror. We devise a modified eight-point algorithm to obtain an initial estimation from 2D joint locations. The 2D joint locations are then refined subject to human body constraints. Finally, a RANSAC algorithm is employed to remove outliers by comparing their epipolar distances to a predetermined threshold. MirrorCalib achieves a rotation error of 1.82° and a translation error of 69.51 mm on a collected real-world dataset, which outperforms the state-of-art method. 
\end{abstract}

\section{Introduction}
To improve health, online coaching and offline learning by watching exercise tutorials are prevalent in daily life. In these tutorials, instructors usually guide students by demonstrating actions in front of a flat mirror in a gym or dance studio. By incorporating a mirrored view of an instructor, both the learning and teaching experience could be enhanced. Especially, instructors can observe and adjust their actions by directly looking at the mirror.

Estimating 3D human motion in exercise videos is crucial to many applications including but not limited to the assessment of trainee exercise quality~\cite{b36}, virtual or augmented reality~\cite{b37}, and motion captioning~\cite{b38, b39}. However, reconstructing 3D human motion from monocular 2D video input is an ill-posed problem, due to depth ambiguity. The inclusion of mirror view in exercise videos provides an additional perspective for the instructors. If the geometric relationship between the real view and the virtual view can be established, in principle, it becomes feasible to determine 3D joint positions through triangulation. 

The procedure for establishing the relationship between two camera views is referred to as camera extrinsic parameter estimation or calibration~\cite{b40}. Extrinsic parameters describe the relative orientation and translation of the camera with respect to a reference frame. Conventionally, the first step in the calibration process involves feature matching of images taken by two cameras. This process includes the detection and association of projections corresponding to the same set of visible features. Feature matching requires an overlapping view of two cameras, which is generally satisfied in stereo systems when the relative rotation between two cameras is small. However, when cameras have little overlapping view (wide baseline) such as a catadioptric system with one real camera and a single mirror (called \textit{virtual camera}) in coaching videos, feature correspondences are difficult to establish.  Even if there exist overlapping views, feature-matching algorithms often fail when the features have complex structures or exhibit repetitive patterns, such as floor or wall tiles~\cite{b41}. 

Fortunately, it is possible to leverage “invisible” features that consistently exist in coaching videos. In coaching videos, an instructor is almost always present in both real and virtual views. Consequently, the joints of the instructor can be exploited as feature correspondences for camera calibration in this context. However, 2D human joint locations detected by the state-of-the-art estimators such as OpenPose~\cite{b25} and HRNet~\cite{b18} are known to be noisy because of occlusion, inconsistent labeling of training data, unbalanced data, etc~\cite{b32, b33}. For instance, it was reported that OpenPose has a 75.6\% mean Average Precision (mAP) on MPII dataset~\cite{b34} and 65.3\% mean average precision over 10 OKS thresholds on COCO validation set~\cite{b35}. Naïve structure-from-motion methods to recover the virtual camera pose all suffer from large estimation noise~\cite{b17}. One key insight is that prior knowledge of human anatomy and motion can be utilized to improve the accuracy of the joint position estimations and thus result in more accurate virtual camera calibration.

\begin{figure*}[t]
    \centering 
    \includegraphics[scale=0.4]{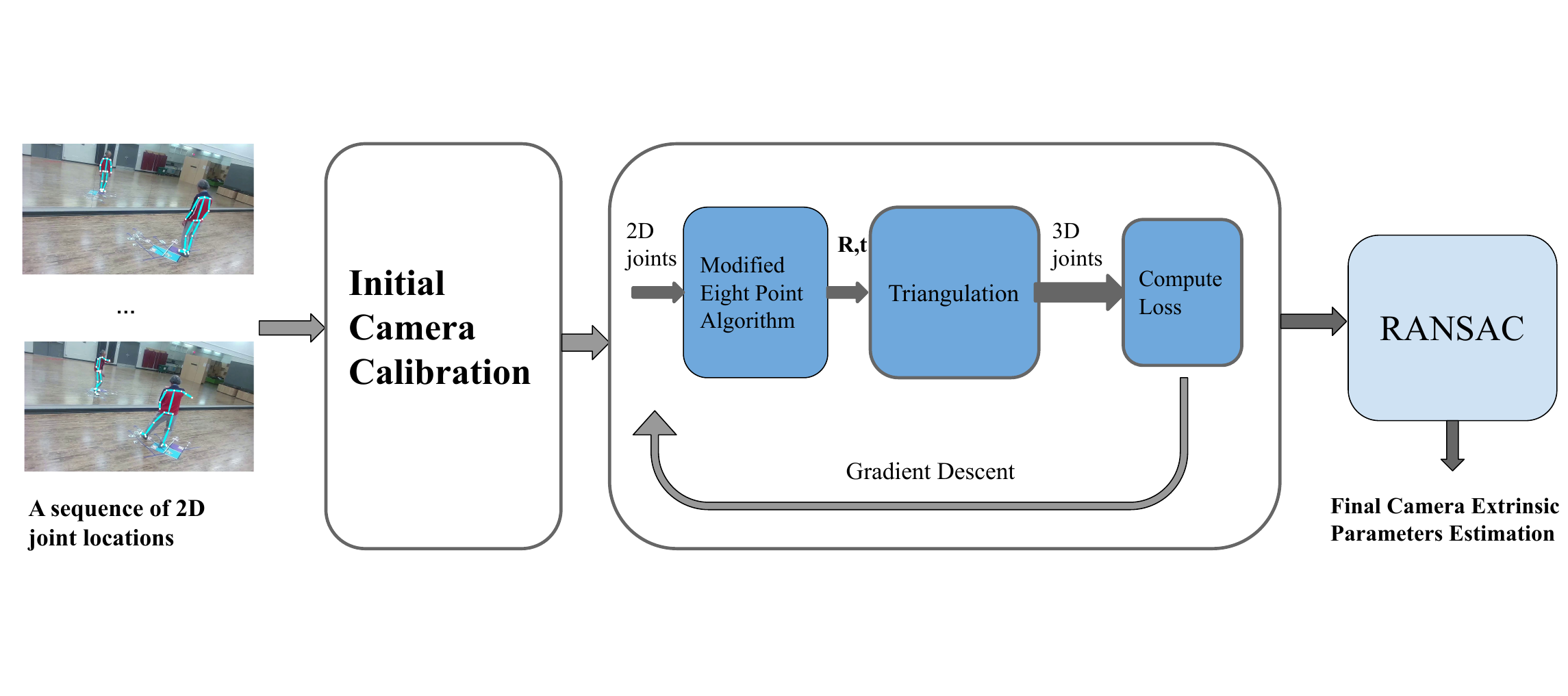}
    \caption{Overview of MirrorCalib. The process takes a video of a human in front of a mirror and passes it through a 2D human pose estimator to obtain the 2D joint locations for both the real and mirrored human. Using a modified eight-point algorithm, an initial estimation of the virtual camera pose is obtained. This estimation is then refined through an optimization process, and a RANSAC algorithm is used to choose the best estimation among all available results.}
    \label{fig: block_diagram}
\end{figure*}

Based on the insight, we propose MirrorCalib, a novel pipeline for estimating the extrinsic parameters of a virtual camera (mirror) relative to a real camera in pre-recorded or streaming exercise videos. MirrorCalib consists of three components as illustrated in Figure \ref{fig: block_diagram}. First, a modified eight-point algorithm with mirror constraints is devised to get the initial camera extrinsic parameters from 2D human joint locations in a sequence of video frames. Next, the 2D joint locations are refined through optimization using characteristics of the human body. After optimization, outlier 2D joint locations are disregarded if their epipolar distances exceed a predefined threshold using random sample consensus (RANSAC). Finally, the extrinsic parameters of the virtual camera with respect to a real camera are calculated from all inliers using the modified eight-point algorithm. The proposed optimization framework and the RANSAC algorithm can be generalized to multi-view setups with real cameras or mirrors.


Compared to existing work that calibrates virtual cameras~\cite{b6, b7, b8, b9}, our work differs in two key aspects. First, previous studies~\cite{b6, b7, b8, b9} often calibrate virtual cameras in catadioptric stereo systems using multiple mirrors. Such systems are not common in daily use. In contrast, MirrorCalib only requires the view of one fixed mirror, which is widely available in gyms and dance studios. Second, since earlier work considers scenarios with more than one mirror, it is easier to obtain overlapping views and rely on visible points to establish correspondences. In contrast, our work utilizes (generally invisible) 2D human joint locations as feature correspondences. Furthermore, although existing work calibrates the extrinsic parameters of real cameras using 2D human joints~\cite{b11, b12}, after initial estimation, extrinsic parameters and 3D human joints are optimized as separate variables. MirrorCalib, in contrast, jointly optimizes the two through the inclusion of epipolar constraints on human joints.

The rest of the paper is organized as follows: a presentation of the necessary background for our approach is provided in Section \ref{background}. In Section \ref{methodology}, a detailed explanation of MirrorCalib is provided. Experiment setup, datasets and evaluation results are presented in Section \ref{evaluation} followed by conclusion in Section \ref{conclusion}.

\section{Background}
\label{background}
In this section, key geometric concepts relevant to estimating virtual camera poses with respect to a real camera are provided. 
\begin{figure}[htbp]
\centerline{\includegraphics[scale = 0.25]{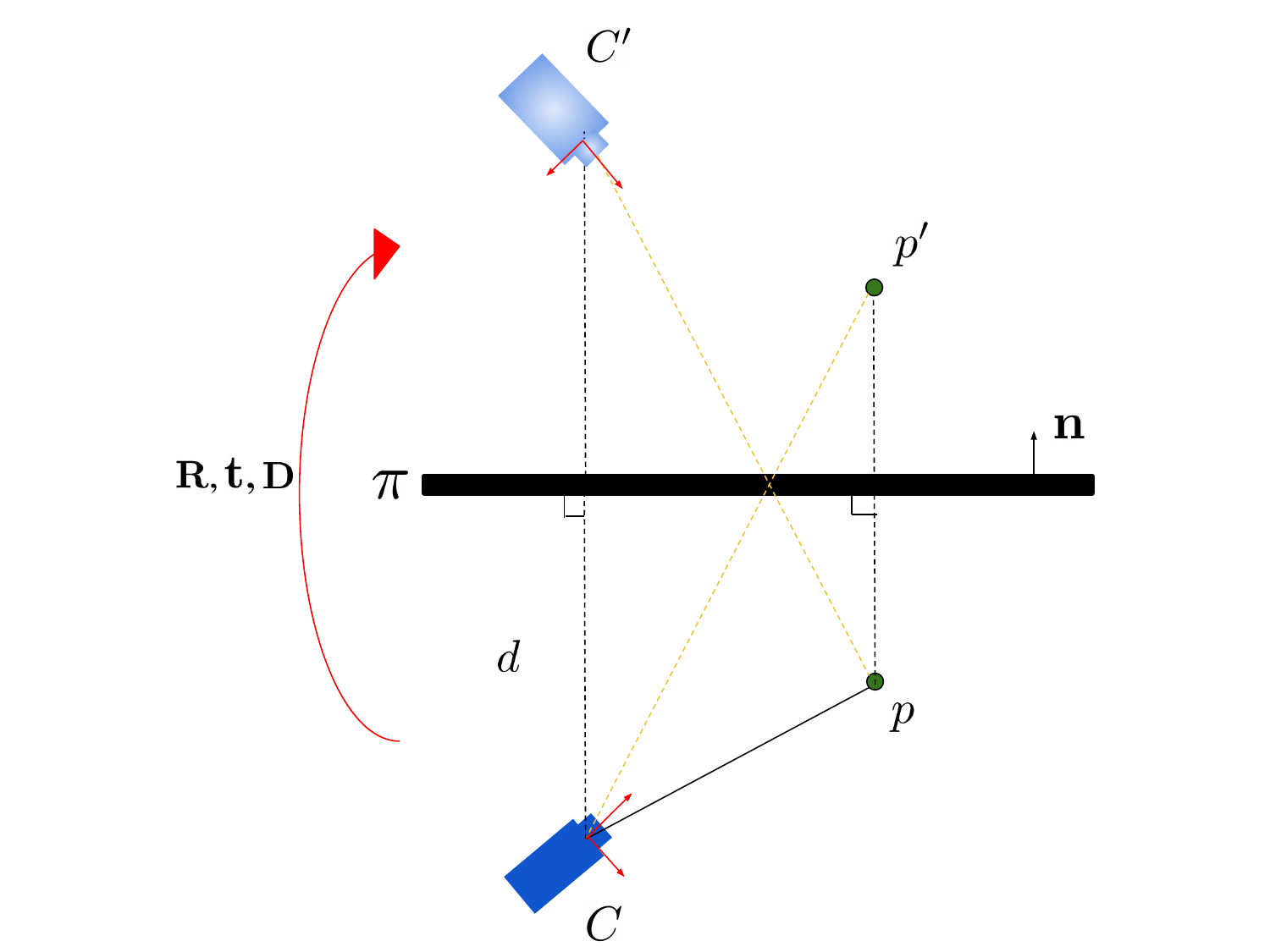}}
\caption{The coordinates frame of real and virtual cameras and their relationship. $\mathbf{R}$, $\mathbf{t}$, $\mathbf{D}$ are rotation, translation and reflection transformation respectively. }
\label{fig_epipolar}
\end{figure}
\subsection{Virtual Camera Projection with Reflective Surfaces and Reflective Epipolar Geometry}
\label{reflective-geo}
In this paper, we treat the real camera frame as the coordinate system where the optical center of the real camera is the origin. Let the intrinsic parameter matrix of the camera be $\mathbf{K}$. We consider the configuration depicted in Figure \ref{fig_epipolar}, in which a perspective camera $C$ is in front of a planar mirror $\pi$ that is uniquely defined by its normal vector $\mathbf{n}$ and a distance $d$ to $C$. 

As shown in Figure \ref{fig_epipolar}, the projection $\mathbf{u^\prime}$ of the mirrored point $p^\prime$ can be seen as the projection $\mathbf{u}''$ of the point $p$ on the image plane of a virtual camera $C'$. The reference frame of the virtual camera is obtained by reflecting that of the real camera about the mirror $\pi$. It should be noted that the real camera is right-handed i.e., the orientation of the camera's coordinate system follows the right-hand rule, while the virtual camera is left-handed. 
Since both the real camera and its virtual counterpart share the same set of intrinsic parameters $\mathbf{K}$, the purpose of camera calibration in this context is to estimate the extrinsic parameters between these two camera frames, specifically, the rotation matrix $\mathbf{R}$ and the translation vector $\mathbf{t}$.

 Denote by $\tilde{\mathbf{u}}$ and $\tilde{\mathbf{u}}'$ the homogeneous coordinates of the point $p$ and the point $p'$ projected onto the image planes of the real camera $C$. From~\cite{b16}, $\tilde{\mathbf{u}}$ and $\tilde{\mathbf{u}}'$ satisfy the reflective epipolar constraint:
\begin{equation}
(\tilde{\mathbf{u}}')^\top\mathbf{F}\tilde{\mathbf{u}} = 0, \label{eq:fundamental}
\end{equation}
where
\begin{equation}
\mathbf{F} \triangleq \mathbf{K}^\top\mathbf{E}\mathbf{K}^{-1}
\end{equation}
is the reflective fundamental matrix, and
\begin{equation}
\mathbf{E} = 2d[\mathbf{n}]_\times 
\end{equation}
is the reflective essential matrix, and $[\mathbf{n}]_\times$ is the skew-symmetric matrix associated with the normal vector $\mathbf{n}$. Note that since the reflective essential matrix is a skew-symmetric matrix, after transformation by the intrinsic camera matrix, the reflective fundamental matrix is also a skew-symmetric matrix.
\\

\section{Methodology}
\label{methodology}
In this section, we present the detailed procedure of MirrorCalib to determine the reflective fundamental matrix from noisy 2D pose estimations. 

\subsection{Eight-Point Algorithm for Solving Reflective Fundamental Matrix}
\label{modified_eight}
 Typically a pair of real cameras have nine unknowns in the fundamental matrix. However, in the case of the reflective fundamental matrix between a real camera and its virtual one, as explained in Section \ref{reflective-geo}, the fundamental matrix is a skew-symmetric matrix determined by only three unknowns. Therefore, the transformed fundamental matrix after image normalization~\cite{b17} is:
 \begin{equation}
\mathbf{F}' = \begin{bmatrix} 0 & x_1 & x_2 \\
-x_1 & 0 & x_3 \\
x_4 & x_5 & x_6 \end{bmatrix},
\end{equation}
where $[x_1, x_2 ... x_6]$ are unknowns. The normalized fundamental matrix can be solved through least-square minimization. Specifically, the matrix $\mathbf{\hat{A}}$ is obtained by forming a row vector for each pair of matching points $[-\hat{v}'\hat{u}+\hat{u}'\hat{v}, \hat{u}', \hat{v}', \hat{u}, \hat{v}, 1]$, where $\hat{u}$'s and $\hat{v}$'s are normalized coordinates. The objective is to minimize:
\begin{equation}
\mathbf{\hat{A}}\mathbf{f'} \label{eq:mo_Af} \quad \text{s.t. } ||\mathbf{f'}|| = 1,
\end{equation}
where $\mathbf{f’}$ is the vector $[x_1, x_2 ... x_6]^\intercal$ that determines $\mathbf{F}'$. The least-square solution to Equation \eqref{eq:mo_Af} can be calculated by getting the smallest eigenvalue of $\mathbf{\hat{A}^\top \hat{A}}$. After $\mathbf{F'}$ is obtained, we transform it to the original reflective fundamental matrix $\mathbf{F}$ by the following Equation \eqref{eq:F'toF}, where $\mathbf{T'}$ and $\mathbf{T}$ are normalization matrix of real view and its virtual view.

\begin{equation}
\mathbf{F} = \mathbf{T'}^\top\mathbf{F'}\mathbf{T},
\label{eq:F'toF}
\end{equation}

\subsection{Reflective Essential Matrix Decomposition}
\label{essential_decomp}
Different from the multi-view epipolar geometry with real cameras, the reference frame of the virtual camera is left-handed. Thus, we decompose the transformation from a real camera reference frame to that of the virtual camera frame into rotation $\mathbf{R}$, translation $\mathbf{t}$, followed by reflection $\mathbf{D}$, as illustrated in Figure \ref{fig_epipolar}. The reflection from the right-handed frame to the left-handed frame can be simply defined as:
\begin{equation}
\mathbf{D} = \begin{bmatrix}
-1 & 0 & 0 \\
0 & 1 & 0 \\
0 & 0 & 1
\end{bmatrix}.
\end{equation}
\hspace{0.422cm} Therefore, the essential matrix that relates the corresponding points on the image planes of the real and virtual cameras is given by:
\begin{equation}
\mathbf{E} = [\mathbf{t'}]_\times \mathbf{D}\mathbf{R} \quad \text{where} \quad  \mathbf{t'} = \mathbf{D}\mathbf{t}.
\end{equation}
\hspace{0.422cm} Similar to a conventional essential matrix, $\mathbf{R}$ and $\mathbf{t}$ can be obtained from the reflective essential matrix $\mathbf{E}$ using singular value decomposition (SVD). 


\subsection{Estimating Virtual Camera Extrinsic Parameters from 2D Human Pose Estimation}
Next, we discuss how to utilize 2D joint coordinates from 2D pose estimation to estimate the extrinsic parameters of the virtual camera $C'$. \\

\subsubsection{Initial Estimation}
\label{initial_estimate}
\hspace{0.422cm}Given the video of a human performing actions in front of a mirror, we first apply a state-of-the-art 2D human pose estimator on each frame to get the 2D joint positions of the real human and the mirrored human. The 2D joint locations on each joint are then matched and used in calibration. Special care must be taken, since the joints of the real human and the mirrored human are symmetrically associated. For instance, the right ankle joint of the real human should correspond to the left ankle of the mirrored human. We employ the modified eight-point algorithm in Section \ref{modified_eight} to calculate the reflective fundamental matrix and the reflective essential matrix. Lastly, the reflective essential matrix is used to decompose and calculate the virtual camera pose with respect to the real camera using singular value decomposition (SVD).

Not all joints are suitable for estimating the virtual camera extrinsic parameters. We omit the joints on one's face, feet, and hands because these areas tend to have a lower resolution and are often occluded in the frames.

\subsubsection{Refining 2D Joint Locations}
\label{sec:optimization}
\hspace{0.422cm}As the key joint positions from 2D human pose estimation tend to be noisy due to occlusion, limited light, and low image resolution, the initial estimates of the extrinsic parameters are inaccurate. To improve their accuracy, we utilize the prior knowledge of the human body. Rather than optimizing the extrinsic parameters directly, we optimize the 2D human joints of the real and virtual human together and use the updated positions to estimate the parameters.

Let $\mathbf{u_{i}^t}$ and $\mathbf{{u_i}^t}'$ denote the 2D coordinates of joint $i$ and reflected joint $i'$ at time $t$, $\mathbf{\Theta}$ and $\mathbf{\Theta'}$ denote the extrinsic parameters of camera $C$ and the virtual camera $C'$ respectively where $\mathbf{\Theta}$ is known. $\mathbf{\Theta'}$ is updated by the modified eight-point algorithm in each iteration, and $\mathbf{X}_i^t$ denote the 3D coordinates of joint $i$ at time $t$. $\mathbf{X}_i^t$ can be calculated through triangulation from $\mathbf{u_{i}^t}$, $\mathbf{{u_i}^t}'$,  $\mathbf{\Theta}$ and $\mathbf{\Theta'}$. Our goal is to ``denoise'' $\mathbf{u}_i^t$ and ${\mathbf{u}_i}^{t'}$, to refine the estimation of $\mathbf{\Theta'}$. 

One body prior is that the length of each limb remains constant over the duration of the recording. Let $l_k^t$ be the length of the $k_{th}$ bone in the bone set $P$ = \{Right Femur, Left Femur, Left Humerus, Right Humerus, Left Ulna, Right Ulna, Left Tibia, Right Tibia, Left Scapular, Right Scaptular, Left Hip, Right Hip\} at time $t$ from estimated 2D joint positions. To account for the discrepancies in bone length, we introduce a penalty term that considers the ratio of bone length variation over time to the mean bone length:
\begin{equation}
L_{\text{var}} = \sum_{k\in P} \frac{\Delta l_k}{\Bar{l_k}},
\label{eq:var}
\end{equation}
where $\Delta l_k$ is the variation of bone length over T frames and $\Bar{l_k}$ is the mean bone length over T frames.

Let $m(\cdot)$ be an operator that maps the index of a bone to the index of its counterpart on the symmetric side of the body. For example, if $k$ is the index of the right femur, $m(k)$ corresponds to the left femur, and vice versa. Under the assumption that the human body is symmetrical, we define a loss term associated with body symmetry as, 
\begin{equation}
L_{\text{sym}} = \sum_t^{T}\sum_{k\in P} ||l_k^t - {l_{m(k)}^t}||.
\label{eq:sym}
\end{equation}
Let $\sigma_k$ be the average normalized length of bone $k$ among abled-bodied humans without physical deformity (eg. according to the statistics from CAESAR Dataset~\cite{b21}). Let $\hat{l_k^t} = \frac{l_k^t}{l_{femur}^t}$, i.e., the length of the $k_{th}$ bone normalized by the length of the longest femur. The loss term associated with anthropometric constraints is thus,  
\begin{equation}
L_{\text{anth}} = \sum_t^{T}\sum_{k\in P}(\hat{l}_k^t - \sigma_k)^2.
\label{eq:anth}
\end{equation}

Let $\mathbf{v_{ml}}^t$ ($\mathbf{v_{mr}}^t$) be the vectors from the left (right) hip to the mid-hip. It is expected that the left hip, right hip, and mid-hip joints are co-linear. Thus, we define:
\begin{equation}
L_{\text{hip}} = \sum_t^{N_t}||{\mathbf{v_{ml}}}^t\times{\mathbf{v_{mr}}}^t||,
\label{eq:hip}
\end{equation}
to penalize abnormal posture of the hips. 

The smoothness of motion is a characteristic of human movement, whereby the change in 3D joint locations between consecutive frames is typically gradual and continuous. Therefore, a loss term associated with the second derivative of 3D joint locations approximated by central finite differences is introduced:

\begin{equation}
L_{\text{smooth}} = \sum_t^{N_t}\sum_{i\in P}^N \mathbf{
\ddot{X}_i^t}.
\end{equation}

Lastly, the denoised 2D joint locations should not deviate too much from the initial estimation. The following reprojection term penalizes the deviation from the original posture:
\begin{equation}
L_{\text{repro}} = \sum_t^{T}\sum_{i\in P}^N \rho(||\Pi(\mathbf{\Theta}, \mathbf{X_i^t}) - \mathbf{u_i^t}|| + ||\Pi(\mathbf{\Theta'}, \mathbf{X_i^t}) - \mathbf{{u_i}^t}'||),
\label{eq:repro}
\end{equation}
where $\Pi$ is the projection operator from 3D to 2D, and $\rho$ denotes the Geman-McClure robust error function to suppress outliers.

To this end, the final objective function can be written by combining Equations \eqref{eq:var} to \eqref{eq:repro} as follows:
\begin{equation}
\begin{aligned}
 &\underset{\text{\textbf{$\mathbf{u,u'}$}}}{\text{min}}\quad \lambda_{\text{var}} L_{\text{var}} +
 \lambda_{\text{sym}} L_{\text{sym}} + \lambda_{\text{anth}} L_{\text{anth}} + \lambda_{\text{hip}} L_{\text{hip}} \\ & + \lambda_{\text{smooth}} L_{\text{smooth}} + \lambda_{\text{repro}} L_{\text{repro}},
\end{aligned}
\label{eq:optimization}
\end{equation}
where $\lambda$'s are empirically chosen weights. 


\subsubsection{Outlier Rejection}

\hspace{0.422cm}After the optimization in Section \ref{sec:optimization}, ``denoised'' 2D joint locations in T frames are obtained. There are many more pairs than needed to solve for the fundamental matrix. To further improve the calibration accuracy, we reject outliers by computing the epipolar distances for eligible pairs for a given fundamental matrix $F$. Let $l(x_i, C') = \mathbf{F}x_i$ and $l(x_i', C) = \mathbf{F}^\top{x_i}'$ be the epipolar lines generated by the 2D point $x_i$ on the virtual view $C'$ and the matched 2D point ${x_i}'$ on the real view $C$ respectively~\cite{Zhang2014}. Then we can compute the sum of the distance from ${x_i}'$ to $l(x_i, C')$ and the distance from $x_i$ to $l({x_i}', C)$ in each frame in a video sequence and denote this sum as $g(\mathbf{F}) = distance(l(x_i, C'), x_i') + distance(l(x_i', C), x_i)$. 

Next, the Random Sample Consensus (RANSAC) algorithm for outlier removal is applied. 
In each iteration, six pairs of 2D joint positions are selected randomly to calculate the fundamental matrix. This matrix then categorizes the pairs into two groups: inliers and outliers, based on their respective epipolar distances. Specifically, the pairs with an epipolar distance larger than a threshold are defined as outliers. The largest inlier set is used to determine the final reflective fundamental matrix as described in Section~\ref{modified_eight}.

\section{Evaluation}
\label{evaluation}
In this section, we evaluate the performance of MirrorCalib on the collected real-world datasets. Additionally, we conduct a case study on 3D human pose estimation to illustrate the practical application of MirrorCalib.
\subsection{Implementation Details}
\subsubsection{Intrinsic Parameters}
\label{eva-intri}
\hspace{0.422cm}MirrorCalib assumes that the intrinsic parameters that characterize the focal length and optical center are given as in previous work~\cite{b11, b12, b13}. These intrinsic parameters can be obtained from the metadata of videos recorded by professional cameras. Alternatively, they can be estimated using vanishing points~\cite{b23} under the assumption that the camera has zero skew, square pixels, and the principal point is located at the image center. 


\subsubsection{2D Human Pose Estimation}
\hspace{0.422cm}In the implementation, we adopt HRNet~\cite{b18} and OpenPose~\cite{b25} to estimate the 2D joint locations of the human and the mirrored human frame by frame from a video sequence of a human moving in front of a mirror. Since HRNet does not provide 2D joint location of the mid-hip, the hip-angle loss is not adopted when HRNet is utilized. The joints except for those on the face and feet are used to obtain an initial estimation of the camera extrinsic parameters. 

\subsubsection{Training}
\hspace{0.422cm}L-BFGS is used in solving the optimization problem in Equation \eqref{eq:optimization}. The learning rate of L-BFGS is set to be 1 and the maximum number of iterations in one optimization step equals 20. The pipeline is implemented in Python.

\subsection{Baseline Methods and Evaluation Metrics}
In order to evaluate the effectiveness of MirrorCalib, we compare its performance with two baseline methods. The first baseline method operates on the same collection of frames and  employs the normalized eight-point algorithm~\cite{b17} without mirror constraints, optimization, or RANSAC. The second baseline method combines the modified eight-point algorithm in initial estimation and optimization using the same strategy in a previous study for multi-view cameras~\cite{b12}\footnote{\scriptsize The method targets uncalibrated and unsynchronized cameras. For a fair comparison, we use synchronized feeds instead.}. Note that there is no prior work on calibrating the extrinsic parameters of a virtual camera relative to a real camera using 2D human joint locations.

To evaluate the accuracy of virtual camera orientations, we calculate the rotation in the axis-angle representation between the ground truth and estimated rotation matrices. The resulting angle is used as the error metric, expressed in degrees. To calculate the translation errors, we transform the estimated translation vector $\mathbf{t}$ so that it has the same scale as the ground-truth translation vector. Then the difference between the ground-truth $\mathbf{\hat{t}}$ and the scaled translation vectors is the translation error expressed in millimeters (mm). The translation error for each estimation is given by:

\begin{equation}
E_t = ||\mu\mathbf{t} - \mathbf{\hat{t}}|| \quad \text{where } \mu = ||\mathbf{\hat{t}}||/||\mathbf{t}||
\end{equation}

\subsection{Datasets}
Since there are no existing video datasets with mirrors including ground truth virtual camera parameters, we collect a real-world dataset specified below.

We recruited seven subjects (5 males and 2 females) performing various actions in front of a mirror and recorded their movements for five minutes per camera location using the WEBCAM C3 model from ASUS. There are a total of 31 different camera positions for these videos. The resolution of those videos is 1920$\cross$1080 and the frame rate is 30 fps. The intrinsic parameters of the camera are obtained by using a chessboard and an OpenCV toolbox~\cite{b24}. To obtain the ground-truth extrinsic parameters, we placed a chessboard in front of the mirror, ensuring that the camera captured both its real and virtual views. After associating the corners of the real and mirrored chessboards, the extrinsic parameters of the virtual camera are calculated using the modified eight-point algorithm and are treated as ground truth.



\begin{figure}[htbp]
\centerline{\includegraphics[scale = 0.10]{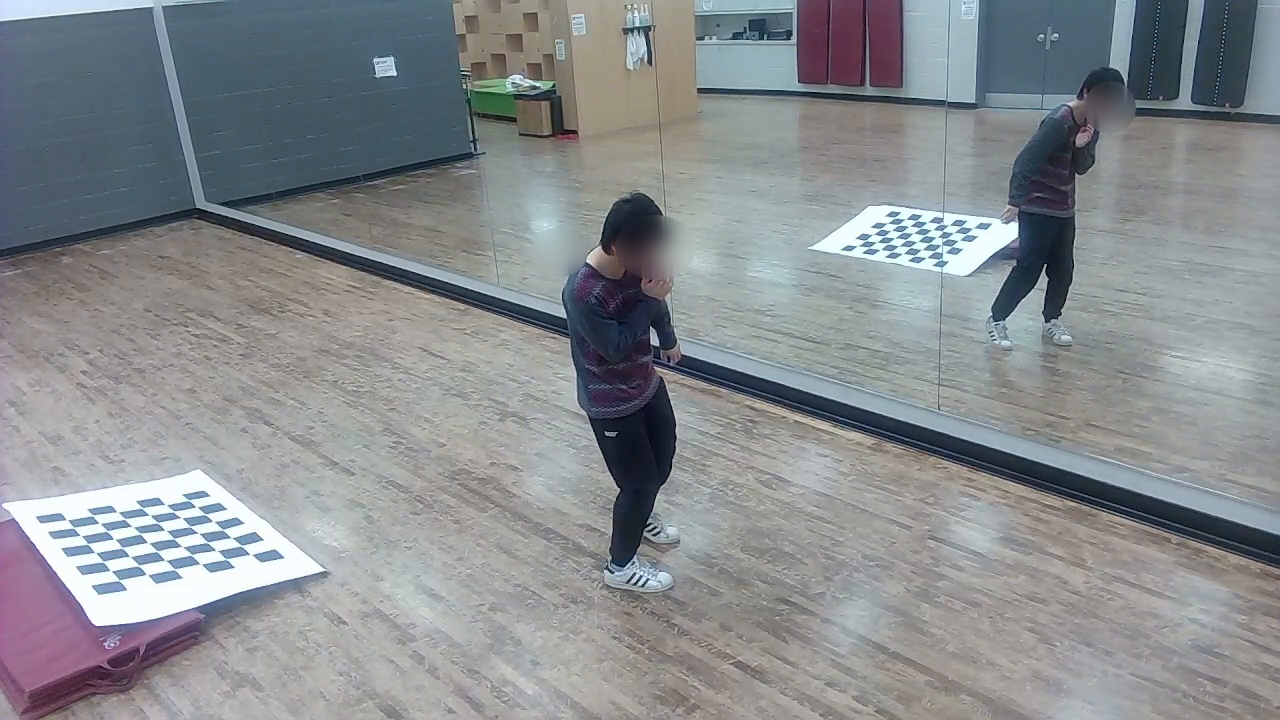}}
\caption{Real dataset collection setup. The chessboard can be captured in both real and virtual views. The corners on the chessboard are only for ground-truth derivation, and are not the key points for MirrorCalib.}
\end{figure}

\subsection{Results}
We conduct experiments on the collected dataset to evaluate the effectiveness of MirrorCalib. In this section, we assume the intrinsic parameters are given. The number of frames T is set to be 1000. Table \ref{tab_real_hrnet} shows the results using 2D joint locations detected by HRNet and OpenPose respectively. For both 2D pose estimators, MirrorCalib outperforms baseline methods. Since HRNet is the most state-of-art method compared to OpenPose, using HRNet as our 2D pose estimator results in higher accuracy in the extrinsic parameter estimation.


\begin{table}[htbp]
\small
\caption{Evaluation of virtual camera pose estimation on real dataset using 2D pose locations detected by HRNet and OpenPose.}.
\begin{center}
\begin{tabular}{|c|c|c|c|}
\hline
\textbf{} & \textbf{} & \textbf{{Translation}} & \textbf{{Rotation}} \\
\textbf{} & \textbf{} & \textbf{{Error (mm)}}& \textbf{{Error (deg)}}\\
\hline
\multirow{3}{*}{HRNet} & Baseline1 & 287.52 $\pm$ 41.57 & 8.91 $\pm$ 2.68 \\
\cline{2-4} 
 &   Baseline2 & 80.21 $\pm$ 25.81 & 2.33 $\pm$ 1.12 \\
 \cline{2-4}
 & \textbf{MirrorCalib} & \textbf{69.51 $\pm$ 19.81} & \textbf{1.82 $\pm$ 0.57} \\
\hline
\multirow{3}{*}{OpenPose} & Baseline1 & 365.28 $\pm$ 58.16 & 12.87 $\pm$ 4.19 \\
\cline{2-4} 
 &   Baseline2 & 116.54 $\pm$ 30.27 & 3.61 $\pm$ 1.83 \\
 \cline{2-4}
 & \textbf{MirrorCalib} & \textbf{91.34 $\pm$ 22.37} & \textbf{2.57$\pm$ 0.69} \\
 
\hline
\end{tabular}
\label{tab_real_hrnet}
\end{center}
\end{table}

\subsection{A Case Study on 3D Human Pose Estimation}
One important application of a catadioptric stereo system is 3D human pose estimation. In the literature, many methods exist that given initial 2D poses from multi-view images, estimate 3D poses in an iterative manner~\cite{b12}. As a case study, we consider a simple pipeline that assumes the intrinsic parameters are known, takes noisy 2D pose estimations as inputs, and calculates 3D joint positions using triangulation using the extrinsic parameters estimated by different methods. We collect a synthetic dataset with ground-truth 3D locations from SMPL~\cite{b19} models of SURREAL dataset~\cite{b27}. To obtain 2D poses, we project the synthetic 3D body joints onto the image planes of both real and virtual cameras using given intrinsic and extrinsic parameters. Gaussian noises are then added to the ground-truth 2D joint locations to obtain noisy 2D joint locations. The mean and standard deviation of noises are set to be 0.076 and 4 pixels. A triangulation algorithm is applied to ground-truth 2D joint locations to obtain the corresponding 3D pose locations using the camera extrinsic parameters from the initial estimation process, the parameters obtained after optimization and RANSAC, and two baseline methods. The performance metric used is PA-MPJPE, which measures the distance (mm) between the predicted 3D key points and their ground truth 3D locations after Procrustes alignment. 

As shown in Table \ref{tab_3d}, 3D pose estimation is significantly more accurate when using the camera extrinsic parameters estimated by MirrorCalib compared to baselines. The result is comparable to SOTA methods for 3D human pose estimation. This experiment also highlights the importance of acquiring precise camera extrinsic parameters for achieving accurate 3D pose estimation. 
\begin{table}[htbp]
\caption{Evaluation of 3D human pose estimation using estimated virtual camera extrinsic parameters.}.
\begin{center}
\begin{tabular}{|c|c|c|c|}
\hline
\textbf{} & \textbf{PA-MPJPE (mm)}\\
\hline
Baseline1 & 168.9 $\pm$  47.3\\
\hline
Baseline2 & 86.4 $\pm$ 30.9\\
\hline
Init & 145.7 $\pm$ 36.5\\
\hline
\textbf{MirrorCalib} & \textbf{68.5 $\pm$ 18.7}\\
\hline
\end{tabular}
\label{tab_3d}
\end{center}
\end{table}

\section{Conclusion}
\label{conclusion}
This paper introduced the novel task of estimating the virtual camera pose relative to the original camera in exercise videos. To achieve this, a modified eight-point algorithm that incorporates mirror characteristics was proposed and the camera parameters were optimized using prior knowledge of the human body. We enhanced the accuracy of our approach with a RANSAC algorithm to reject all outliers. The effectiveness of our method was demonstrated on a collected real-world dataset and a case study of 3D human pose estimation was also provided to confirm its practical applicability. The optimization process and RANSAC algorithm have the potential to be extended to general multi-view camera pose estimation with humans present in the scene. In future work, we will focus on improving the computational efficiency of MirrorCalib.


\begin{thebibliography}{1}
\small
\bibliographystyle{ieee}

\bibitem{b1} R. Rodrigues, J. P. Barreto, and U. Nunes, “Camera pose estimation using images of Planar Mirror Reflections,” Computer Vision – ECCV 2010, pp. 382–395, 2010. 
\bibitem{b2} K. Takahashi, S. Nobuhara, and T. Matsuyama, “A new mirror-based extrinsic camera calibration using an orthogonality constraint,” in Proceedings of the IEEE Computer Society Conference on Computer Vision and Pattern Recognition, 2012.
\bibitem{b3} R. K. Kumar, A. Ilie, J. M. Frahm, and M. Pollefeys, “Simple calibration of non-overlapping cameras with a mirror,” in 26th IEEE Conference on Computer Vision and Pattern Recognition, CVPR, 2008. 
\bibitem{b4} G. Long, L. Kneip, X. Li, X. Zhang, and Q. Yu, “Simplified mirror-based camera pose computation via rotation averaging,” in Proceedings of the IEEE Computer Society Conference on Computer Vision and Pattern Recognition, 2015, vol. 07-12-June-2015.
\bibitem{b5} A. Agrawal and S. Ramalingam, “Single image calibration of multi-axial imaging systems,” in Proceedings of the IEEE Computer Society Conference on Computer Vision and Pattern Recognition, 2013. doi: 10.1109/CVPR.2013.184.
\bibitem{b6} S. A. Nene and S. K. Nayar, “Stereo with mirrors,” in Proceedings of the IEEE International Conference on Computer Vision, 1998. doi: 10.1109/iccv.1998.710852.
\bibitem{b7} J. Gluckman and S. K. Nayar, “Catadioptric stereo using planar mirrors,” Int J Comput Vis, vol. 44, no. 1, 2001, doi: 10.1023/A:1011172403203.
\bibitem{b8} T. Tahara, R. Kawahara, S. Nobuhara, and T. Matsuyama, “Interference-Free Epipole-Centered Structured Light Pattern for Mirror-Based Multi-view Active Stereo,” in Proceedings - 2015 International Conference on 3D Vision, 3DV 2015, 2015. doi: 10.1109/3DV.2015.25.
\bibitem{b9} D. Lanman, D. Crispell, and G. Taubin, “Surround structured lighting: 3-D scanning with orthographic illumination,” Computer Vision and Image Understanding, vol. 113, no. 11, 2009, doi: 10.1016/j.cviu.2009.03.016.
\bibitem{b10} X. Ying, K. Peng, Y. Hou, S. Guan, J. Kong, and H. Zha, “Self-calibration of catadioptric camera with two planar mirrors from silhouettes,” IEEE Trans Pattern Anal Mach Intell, vol. 35, no. 5, 2013, doi: 10.1109/TPAMI.2012.195.
\bibitem{b11} J. Puwein, L. Ballan, R. Ziegler, and M. Pollefeys, “Joint Camera Pose Estimation and 3D human pose estimation in a multi-camera setup,” Computer Vision -- ACCV 2014, pp. 473–487, 2015. 
\bibitem{b12} K. Takahashi, D. Mikami, M. Isogawa, and H. Kimata, “Human pose as calibration pattern: 3D human pose estimation with multiple unsynchronized and uncalibrated cameras,” in IEEE Computer Society Conference on Computer Vision and Pattern Recognition Workshops, 2018, vol. 2018-June. doi: 10.1109/CVPRW.2018.00230.
\bibitem{b13} B. Huang, Y. Shu, T. Zhang, and Y. Wang, “Dynamic Multi-Person Mesh Recovery from Uncalibrated Multi-View Cameras,” in Proceedings - 2021 International Conference on 3D Vision, 3DV 2021, 2021. doi: 10.1109/3DV53792.2021.00080.
\bibitem{b14} G. Ben-Artzi, Y. Kasten, S. Peleg, and M. Werman, “Camera Calibration from Dynamic Silhouettes Using Motion Barcodes,” in Proceedings of the IEEE Computer Society Conference on Computer Vision and Pattern Recognition, 2016, vol. 2016-December. doi: 10.1109/CVPR.2016.444.
\bibitem{b15} S. N. Sinha and M. Pollefeys, “Camera network calibration and synchronization from silhouettes in archived video,” Int J Comput Vis, vol. 87, no. 3, 2010, doi: 10.1007/s11263-009-0269-2.
\bibitem{b16} G. L. Mariottini, S. Scheggi, F. Morbidi, and D. Prattichizzo, “Planar mirrors for image-based robot localization and 3-D reconstruction,” Mechatronics, vol. 22, no. 4, 2012
\bibitem{b17} R. I. Hartley, “In defence of the 8-point algorithm,” in IEEE International Conference on Computer Vision, 1995.
\bibitem{b18} J. Wang et al., “Deep High-Resolution Representation Learning for Visual Recognition,” IEEE Trans Pattern Anal Mach Intell, vol. 43, no. 10, 2021.
\bibitem{b19} M. Loper, N. Mahmood, J. Romero, G. Pons-Moll, and M. J. Black, “SMPL: A skinned multi-person linear model,” in ACM Transactions on Graphics, 2015, vol. 34, no. 6.
\bibitem{b20} C. Ionescu, D. Papava, V. Olaru, and C. Sminchisescu, “Human3. 6M,” IEEE Transactions on Pattern Analysis and Machine Intelligence, 2014.

\bibitem{b21} K. M. Robinette, H. Daanen, and E. Paquet, “The CAESAR project: A 3-D surface anthropometry survey,” in Proceedings - 2nd International Conference on 3-D Digital Imaging and Modeling, 3DIM 1999, 1999. doi: 10.1109/IM.1999.805368.
\bibitem{b22} R. Y. Tsai and T. S. Huang, “Uniqueness and Estimation of Three-Dimensional Motion Parameters of Rigid Objects with Curved Surfaces,” IEEE Trans Pattern Anal Mach Intell, vol. PAMI-6, no. 1, 1984, doi: 10.1109/TPAMI.1984.4767471.
\bibitem{b23} Q. Fang, Q. Shuai, J. Dong, H. Bao, and X. Zhou, “Reconstructing 3D human pose by watching humans in the mirror,” in Proceedings of the IEEE Computer Society Conference on Computer Vision and Pattern Recognition, 2021. doi: 10.1109/CVPR46437.2021.01262.
\bibitem{b24} G. Bradski, “The OpenCV Library,” Dr. Dobb’s Journal of Software Tools, 2000.
\bibitem{b25} Z. Cao, G. Hidalgo, T. Simon, S.-E. Wei, and Y. Sheikh, “OpenPose: Realtime multi-person 2D pose estimation using part affinity fields,” IEEE Transactions on Pattern Analysis and Machine Intelligence, vol. 43, no. 1, pp. 172–186, 2021. 
\bibitem{b26} P. Sturm and T. Bonfort, “How to compute the pose of an object without a direct view?,” Computer Vision – ACCV 2006, pp. 21–31, 2006. 
\bibitem{b27} G. Varol et al., “Learning from synthetic humans,” in Proceedings - 30th IEEE Conference on Computer Vision and Pattern Recognition, CVPR 2017, 2017. doi: 10.1109/CVPR.2017.492.
\bibitem{b28} K. H. Jang, Y. M. Cha, and S. K. Jung, “3D reconstruction using moving planar mirror,” in IASTED International Conference on Computer Graphics and Imaging, 2003.
\bibitem{b29} M. Kanbara, N. Ukita, M. Kidode, and N. Yokoya, “3D scene reconstruction from reflection images in a spherical mirror,” in Proceedings - International Conference on Pattern Recognition, 2006. doi: 10.1109/ICPR.2006.32.
\bibitem{b30} H. Zhong, W. F. Sze, and Y. S. Hung, “Reconstruction from plane mirror reflection,” in Proceedings - International Conference on Pattern Recognition, 2006. doi: 10.1109/ICPR.2006.981.
\bibitem{b31} B. Hu, “It’s all done with mirrors: calibration- and- correspondence- free 3D reconstruction,” in Proceedings of the 2009 Canadian Conference on Computer and Robot Vision, CRV 2009, 2009. doi: 10.1109/CRV.2009.29.
\bibitem{b32} Y. Zhang, C. Wang, X. Wang, W. Liu, and W. Zeng, “VoxelTrack: Multi-Person 3D Human Pose Estimation and Tracking in the Wild,” IEEE Trans Pattern Anal Mach Intell, vol. 45, no. 2, 2023, doi: 10.1109/TPAMI.2022.3163709.
\bibitem{b33} C. Lassner, J. Romero, M. Kiefel, F. Bogo, M. J. Black, and P. V. Gehler, “Unite the people: Closing the loop between 3D and 2D human representations,” in Proceedings - 30th IEEE Conference on Computer Vision and Pattern Recognition, CVPR 2017, 2017. doi: 10.1109/CVPR.2017.500.
\bibitem{b34} M. Andriluka, L. Pishchulin, P. Gehler, and B. Schiele, “2D human pose estimation: New benchmark and state of the art analysis,” in Proceedings of the IEEE Computer Society Conference on Computer Vision and Pattern Recognition, 2014. doi: 10.1109/CVPR.2014.471.
\bibitem{b35} T.-Y. Lin et al., “Microsoft COCO: Common Objects in Context,” Proceedings of the IEEE Computer Society Conference on Computer Vision and Pattern Recognition, 2015.
\bibitem{b36} M. Fieraru, M. Zanfir, S. C. Pirlea, V. Olaru, and C. Sminchisescu, “AIFit: Automatic 3D Human-Interpretable Feedback Models for Fitness Training,” in Proceedings of the IEEE Computer Society Conference on Computer Vision and Pattern Recognition, 2021. doi: 10.1109/CVPR46437.2021.00979.
\bibitem{b37} C. Sminchisescu, “3D Human motion analysis in monocular video techniques and challenges,” in Proceedings - IEEE International Conference on Video and Signal Based Surveillance 2006, AVSS 2006, 2006. doi: 10.1109/AVSS.2006.3.
\bibitem{b38} Y. Goutsu and T. Inamura, “Linguistic Descriptions of Human Motion with Generative Adversarial Seq2Seq Learning,” in Proceedings - IEEE International Conference on Robotics and Automation, 2021. doi: 10.1109/ICRA48506.2021.9561519.
\bibitem{b39} W. Takano and Y. Nakamura, “Statistical mutual conversion between whole body motion primitives and linguistic sentences for human motions,” International Journal of Robotics Research, vol. 34, no. 10, 2015, doi: 10.1177/0278364915587923.
\bibitem{b40} S. J. D. Prince, Computer Vision: Models, Learning, and Inference. New York: Cambridge University Press, 2014. 
\bibitem{b41} S. Beckouche, S. Leprince, N. Sabater, and F. Ayoub, “Robust outliers detection in image point matching,” in Proceedings of the IEEE International Conference on Computer Vision, 2011. doi: 10.1109/ICCVW.2011.6130241.
\end{thebibliography}
\end{document}